\documentclass[12pt]{article}
\usepackage[english]{babel}
\usepackage[utf8]{inputenc}
\usepackage{johd}
\usepackage{fancyhdr}
\fancypagestyle{preprint}{%
  \fancyhf{}%
  \fancyhead[L]{\small Preprint. Under review.\vspace{2pt}\hrule}%
  \fancyfoot[C]{\thepage}%
}
\usepackage{microtype}
\usepackage{booktabs}
\usepackage{todonotes}
\usepackage{soul}
\usepackage{tcolorbox}
\tcbuselibrary{breakable}
\usepackage{enumitem}
\usepackage{amssymb}
\usepackage{graphicx}

\newcommand{\pmark}{%
  \rlap{\checkmark}\kern0.3em\raisebox{1.1ex}{\rotatebox{-45}{\rule{0.5em}{0.4pt}}}%
}
\newcommand{\cmark}{\checkmark}
\newcommand{\xmark}{$\times$}

\newtcolorbox{promptbox}{
  colback=gray!5,
  colframe=gray!50,
  boxrule=0.5pt,
  arc=2pt,
  left=6pt,
  right=6pt,
  top=6pt,
  bottom=6pt,
  breakable
}

\definecolor{darkblue}{rgb}{0, 0, 0.5}
\hypersetup{colorlinks=true, citecolor=darkblue, linkcolor=darkblue, urlcolor=darkblue, hypertexnames=false}

\newcounter{algorithm}

\title{AutoSurfer -- Teaching Web Agents through Comprehensive Surfing, Learning, and Modeling}

\author{Fazle Elahi Faisal$^{*}$, Qianhui Wu, Baolin Peng, Jianfeng Gao \\
        \small Microsoft Research \\\\
        \small $^{*}$Corresponding author: \tt{fafaisal@microsoft.com} \\
}
\date{}

\begin{document}

\maketitle
\thispagestyle{preprint}

\begin{abstract}
Recent advances in multimodal large language models (LLMs) have revolutionlized web agents that can automate complex tasks on websites. However, their accuracy remains limited by the scarcity of high-quality web trajectory training data. Existing automatic trajectory generation methods suffer from incomplete website coverage due to homepage-based task proposals or random-walk exploration. Such methods often result in hallucinated or ambiguous task synthesis that lead to incomplete and unreliable trajectory generation. Here, we present \textit{AutoSurfer}, a comprehensive web trajectory generator that addresses these limitations through three key innovations. First, AutoSurfer employs a systematic breadth-first exploration strategy that maintains a queue of discovered pages and action traces, propagates knowledge across pages to avoid redundant exploration, and recursively expands multi-level graphical user interface (GUI) elements---closely resembling how a human would learn a new website. Second, AutoSurfer leverages the exploration trajectory to guide task synthesis, reducing hallucinations by grounding complex tasks in actual navigation paths rather than isolated actions or page content alone. Third, AutoSurfer uses the same exploration trajectory as hints to steer a web agent toward more accurate and reliable trajectory refinement. Together, these innovations enable AutoSurfer to comprehensively cover a website's action space and generate data suitable for training website-specific LLMs (wLLMs). We evaluate AutoSurfer on the WebArena benchmark by fine-tuning Qwen2.5-VL-7B-Instruct and demonstrate that it outperforms state-of-the-art methods---Explorer, OS-Genesis, and SynthAgent---achieving up to 24.23\% overall task completion accuracy compared to 19.59\% for the best prior method. Further, task diversity analysis demonstrates that AutoSurfer yields a more diverse distribution of synthesized tasks than prior approaches.
\end{abstract}

\section{Introduction}
\label{sec:intro}

Modern websites are becoming increasingly complex, which raises the cognitive and operational burden on users and highlights the need for web task automation.
Recent advancements in multimodal LLMs have enabled agentic AI systems that can automate (or semi-automate) web task execution \citep{su2024Language, Chen2024InternVL, wang2024qwen2vl, sodhi2024stepstackedllmpolicies, fourney2024magentic, yang2024agentoccamsimplestrongbaseline, zhang2025symbioticcooperationwebagents, openai2024cua}. 
The first phase of development focused on copilots, which help users gather and synthesize information from one or more websites based on their goals and constraints \citep{Nath2025ActionEngine}. For example, a travel-planning copilot can compile and recommend itineraries from a user request, but the user must still manually complete the required bookings. 
The second phase focused on web agents, which address this limitation by automating task execution on the web. 
These web agents extend the capabilities of copilots by automatically executing web tasks from natural-language instructions. For example, an agent can not only generate a travel plan but also complete the necessary bookings on the user's behalf. Notable web agents include SteP \citep{sodhi2024stepstackedllmpolicies}, Magentic-One \citep{fourney2024magentic}, AgentOccam \citep{yang2024agentoccamsimplestrongbaseline}, AgentSymbiotic \citep{zhang2025symbioticcooperationwebagents}, and CUA \citep{openai2024cua}.

However, web agents remain far from widespread adoption due to challenges in accuracy, speed, and privacy. 
One major reason for their limited accuracy is the lack of high-quality web interaction data for training popular and widely used LLMs \citep{koh2024visualwebarena}. 
Consequently, when an agent attempts a complex task, the underlying model may fail to translate user intent into the correct sequence of GUI actions \citep{koh2024visualwebarena}. 
This challenge is even more pronounced on previously unseen websites, where agents often fail to anticipate deeper functionality and therefore struggle to complete multi-step workflows \citep{koh2024visualwebarena}. 
While newer and more powerful LLMs can improve accuracy \citep{yuan2026efficientllm}, they often come at the cost of higher latency and expense. 
Recent work therefore emphasizes smaller yet high-quality models that preserve accuracy while improving efficiency and reducing cost \citep{sun2025osgenesisautomatingguiagent,wang2026adaptingwebagentssynthetic,awadallah2025fara7b}. The on-device viability of such smaller LLMs can also strengthen privacy and security by keeping user data local \citep{awadallah2025fara7b}.

High-quality training data that captures realistic tasks across diverse websites and their corresponding trajectories is essential for training high-quality web agents. Early approaches relied mainly on human annotation, which is expensive and time-consuming \citep{li2024on,lu2024weblinx}. Later methods, such as AgentTrek \citep{xu2025agenttrek}, leveraged website tutorials to synthesize tasks and generate trajectories through LLM reasoning. However, tutorials are often limited and incomplete, and tutorial-based synthesis can suffer from hallucination and weak alignment with user intent \citep{wang2026adaptingwebagentssynthetic}. More recent interaction-driven methods improve task synthesis by exploring websites directly and deriving tasks from those interactions. For example, Explorer \citep{pahuja2025explorer} proposes tasks from a website's homepage and iteratively refines them through further interactions, but this homepage-centered strategy can overlook deeper site functionality and therefore miss important tasks. OS-Genesis \citep{sun2025osgenesisautomatingguiagent} instead combines random walk-based exploration with reverse task synthesis to discover tasks and then uses a web agent to refine them. However, its random exploration can miss many GUI actions, and the discovered tasks can remain hallucinated or ambiguous because they are inferred from local and isolated web interaction analysis, which in turn weakens trajectory generation. SynthAgent \citep{wang2026adaptingwebagentssynthetic} improves the refinement stage and yields more reliable trajectories, but it inherits the same exploration bottlenecks as OS-Genesis, including missed functionality and noisy task discovery. A broader discussion on prior methods are elaborated in Section \ref{sec:relatedwork}.

\begin{figure}[t]
    \centering
    \includegraphics[width=0.9\linewidth]{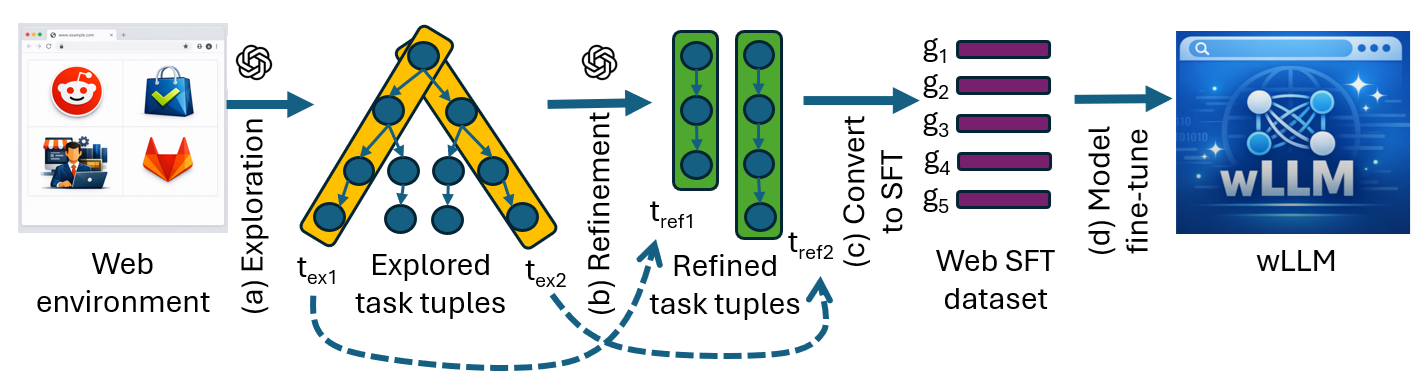}
  \caption{The full architecture of AutoSurfer. Given a web environment, (a) AutoSurfer comprehensively explores the website to discover complex tasks and their trajectories, (b) AutoSurfer refines the discovered tasks and trajectories to improve correctness and quality while reducing hallucinations, (c) AutoSurfer converts the refined tasks and trajectories into an SFT dataset, and (d) AutoSurfer uses the SFT dataset to train a wLLM.}
    \label{fig:method:autosurfer}
\end{figure}

\begin{figure*}[t]
\centering
\begin{minipage}[t]{0.51\textwidth}
\vspace{0pt}
\centering
\refstepcounter{table}\label{tab:comparison}
\small
\resizebox{\linewidth}{!}{%
\begin{tabular}{lccccc}
\toprule
\textbf{Property} & \textbf{Agent} & \textbf{OS-} & \textbf{Explorer} & \textbf{Synth} & \textbf{Auto} \\
 & \textbf{Trek} & \textbf{Genesis} &  & \textbf{Agent} & \textbf{Surfer} \\
\midrule
Systematic & \xmark & \xmark & \xmark & \pmark & \cmark \\
traversal &  &  &  &  & \\
\cmidrule(lr){1-6}
Menu/dropdown  & \xmark & \xmark & \xmark & \xmark & \cmark \\
handling &  &  &  &  & \\
\cmidrule(lr){1-6}
Fixed vs.  & \xmark & \xmark & \xmark & \pmark & \cmark \\
dynamic elements &  &  &  &  & \\
\cmidrule(lr){1-6}
Coupled  & \xmark & \xmark & \xmark & \pmark & \cmark \\
task-trajectory &  &  &  &  & \\
\cmidrule(lr){1-6}
Cross-page  & \xmark & \xmark & \xmark & \xmark & \cmark \\
deduplication &  &  &  &  & \\
\cmidrule(lr){1-6}
Task  & \xmark & \cmark & \cmark & \cmark & \cmark \\
refinement &  &  &  &  & \\
\cmidrule(lr){1-6}
Trajectory  & \xmark & \xmark & \xmark & \cmark & \cmark \\
refinement &  &  &  &  & \\
\bottomrule
\end{tabular}%
}

\vspace{0.4em}
{\small\textbf{Table \thetable.} Comparison of AutoSurfer with other state-of-the-art trajectory generators -- AgentTrek, OS-Genesis, Explorer, and SynthAgent.}
\end{minipage}
\hfill
\begin{minipage}[t]{0.46\textwidth}
\vspace{0pt}
\centering
\includegraphics[width=\linewidth]{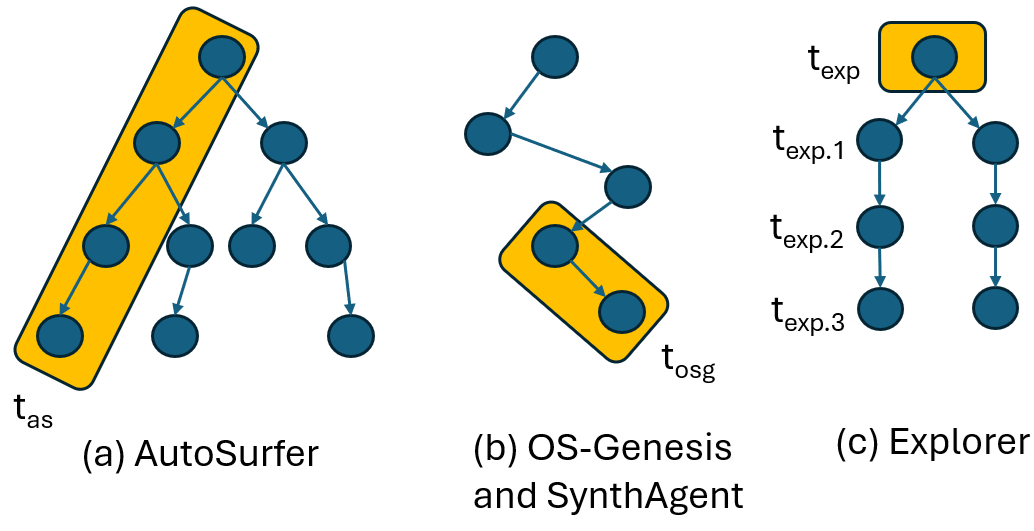}

\refstepcounter{figure}\label{fig:method:tasksynthesis}
\vspace{0.4em}
{\small\textbf{Figure \thefigure.} Illustration of task synthesis by different methods. AutoSurfer, OS-Genesis/SynthAgent, and Explorer synthesizes a task based on (a) full exploration trajectory, (b) last action only, and (c) homepage content only (with refinements in subsequent pages), respectively.}
\end{minipage}
\end{figure*}

In this paper, we present \textit{AutoSurfer} (Figure \ref{fig:method:autosurfer}), a novel and comprehensive web trajectory generator, that addresses the limitations of existing methods in terms of exploration, task discovery and synthesis, and trajectory generation (Table \ref{tab:comparison}). Further, we demonstrate that AutoSurfer-generated data can meaningfully train a website-specific LLM (wLLM). To best of our knowledge, no prior works are explicitly designed to train a wLLM. Key innovations by AutoSurfer are summarized below.

\begin{itemize}
  \item \textbf{Systematic and comprehensive exploration}. AutoSurfer uses a novel, systematic and comprehensive exploration strategy that addresses the limited website coverage by task proposal-based methods (such as AgentTrek and Explorer) as well as random walk-based methods (such as OS-Genesis and SynthAgent). AutoSurfer explores a website in a breadth-first manner by maintaining a queue of discovered pages and their corresponding action traces. AutoSurfer learns a website gradually during exploration by passing knowledge from prior pages to subsequent pages. In addition, AutoSurfer uses the propagated knowledge to avoid redundant exploration. Crucially, AutoSurfer treats static and dynamic elements differently: it systematically traverses static functionality, including recursively expanding menu and dropdown options, while selectively sampling dynamic content to capture representative behaviors without redundant exploration. 
  \item \textbf{Task synthesis with exploration trajectory}. AutoSurfer's systematic breadth-first exploration produces meaningful trajectories that preserve the sequence of intermediate states and actions leading to a discovered capability. AutoSurfer uses these exploration trajectories to augment existing task synthesis approaches, allowing synthesized tasks to be grounded not only in isolated page content or a final action, but also in the broader navigation context through which the task emerges. This design helps capture user intent more accurately, improves the coherence of synthesized tasks, and reduces hallucinations. To the best of our knowledge, prior work has not explicitly used exploration trajectories to guide task synthesis in this way.
  \item \textbf{Trajectory generation with exploration trajectory}. In addition to task synthesis, AutoSurfer leverages the exploration trajectory to guide a web agent toward more accurate and reliable trajectory generation. Prior methods rely solely on the LLM reasoning to refine tasks and generate trajectories, which can easily drift toward an incorrect execution path, especially when the discovered task is hallucinated or ambiguous. In contrast, AutoSurfer uses the exploration trajectory as an additional source of grounding that complements the LLM reasoning and steers the web agent toward the correct direction.
  \item \textbf{Training wLLM}. Unlike previous approaches, AutoSurfer aims to understand an entire website and generate tasks and trajectories that maximize coverage of the website's action space and functionality. This distinctive property makes AutoSurfer-generated data well suited for training a website-specific LLM (wLLM) that captures the website's full capability. In contrast, prior task proposal-based and random walk-based methods are designed to collect only a limited set of representative task and trajectory samples from a website, which is generally insufficient for training a wLLM. To the best of our knowledge, AutoSurfer is the first work explicitly designed to train a wLLM.
\end{itemize}

\section{Method}
\label{sec:method}

AutoSurfer first explores the environment to discover potential tasks and their trajectories (i.e., task-trajectory pairs, or task tuples; Section \ref{sec:method:problem}) using a novel breadth-first traversal strategy (Section \ref{sec:method:exploration:traversal}). During this exploration process, AutoSurfer synthesizes complex tasks by combining simple tasks that are logically connected (Section \ref{sec:method:exploration:tasksynthesis}). It then refines the discovered complex task tuples to improve correctness and quality while reducing hallucinations (Section \ref{sec:method:refinement}). Finally, AutoSurfer converts the refined task tuples into a supervised fine-tuning (SFT) dataset and uses the dataset to train an LLM (Section \ref{sec:method:finetuning}).

\subsection{Problem Formulation}
\label{sec:method:problem}

Given a website $W$, $T_W$ is the set of all possible tasks that can be performed on the website. Each task $t \in T_W$ can be accomplished by performing a sequence of GUI actions $A_t = \{a_1, a_2, ..., a_n\}$ on $W$, where each action $a_i$ corresponds to a specific GUI action from a predefined action space $A_{all} = \{a_{click}, a_{fill}, ...\}$ that can be performed on the website. Each action $a_i$ is determined based on a task description  $td$ and an observation $o_i$. The trajectory of a task $t$ is defined as a sequence of observation-action pairs and a final observation, represented as $tr_t = \{o_1, a_1, o_2, a_2, ..., o_{n-1}, a_{n-1}, o_n\}$. Therefore, each task $t$ can be denoted as a tuple of task description and trajectory, represented as $t = (td_t, tr_t)$. The goal of AutoSurfer is to comprehensively explore $W$ to discover maximum possible tasks $T_W^{AS} \subseteq T_W$ so that $|T_W - T_W^{AS}|$ is minimized. For simplicity, we use the term `task tuple' to refer to a task description and its trajectory, and `task' to refer to the task description alone in the rest of the paper. We define complex task as a task that requires at least two GUI actions to complete. In this paper, we use `task' and `complex task' interchangeably.

\subsection{Exploration}
\label{sec:method:exploration}

\subsubsection{Traveral}
\label{sec:method:exploration:traversal}

Unlike the random walk-based traversal used in most prior work \citep{sun2025osgenesisautomatingguiagent,wang2026adaptingwebagentssynthetic}, AutoSurfer explores a website using a systematic breadth-first strategy. It maintains a queue of visited pages, initialized with a seed page, typically the homepage, and uses LLM reasoning to discover simple tasks on each page (Appendix \ref{appendix:taskdiscoveryprompt}). We define a simple task as one that can be completed using GUI elements on the same page, ranging from single-step actions such as `navigate to Forums page' to multi-step actions such as `search for apartment recommendation at Pittsburgh'. We denote the initially discovered simple tasks as $\mathcal{S}_{init}$ (Algorithm~\ref{alg:exploration_task_tuple_generation}).
AutoSurfer then removes redundant, risky and unimportant tasks from $\mathcal{S}_{init}$, and then adds information-seeking tasks (described below) to form the final set of simple tasks, denoted as $\mathcal{S}$ (Algorithm~\ref{alg:exploration_task_tuple_generation}). It executes these tasks one by one; if execution reaches a new page, that page and its action trace are enqueued for further exploration. Exploration stops when all discovered pages have been explored or when the number of synthesized complex tasks reaches a preset limit of 1,000.

\noindent\textbf{Traverse Menu and Dropdown Elements.} Menu and dropdown options are often hidden until expanded, so AutoSurfer cannot identify them from the initial page view alone. For example, clicking the profile name in Reddit reveals links to the profile page, settings page, and related options. AutoSurfer therefore begins with tasks covering visible elements, which may include a task such as `expand the profile menu'. After executing each simple task, it checks whether the page remains the same but shows a moderate visual change, using LLM reasoning (Appendix \ref{appendix:visualchangeprompt}); this indicates that a menu or dropdown has likely been expanded. AutoSurfer then identifies newly visible elements and discovers additional simple tasks from only those new elements (Appendix \ref{appendix:differentialtaskdiscoveryprompt}), denoted as $\mathcal{S}_{new}$ in Algorithm~\ref{alg:exploration_task_tuple_generation}. These new tasks extend the parent expansion task into a hierarchy (Figure \ref{fig:method:traversal}, middle panel), and AutoSurfer recursively repeats the process for nested expandable elements. Once the full hierarchy is uncovered, AutoSurfer removes intermediate expansion-only tasks and retains only the leaf tasks, while reconstructing each leaf task with the full action prefix from its ancestors. For instance, `navigate to user settings' becomes a two-step task: click the profile name, then click the settings option. This allows AutoSurfer to systematically traverse multi-level menus and dropdowns without prior knowledge of their structure.


\noindent\textbf{Avoid Repetitive Elements Across Pages.} Many websites repeat the same functionality across multiple pages. For example, Reddit exposes search on the home page, forums page, and many other pages, so discovering it once is usually sufficient; tasks that require search can be executed from the home page without rediscovering the same capability elsewhere. AutoSurfer therefore uses its queue to propagate previously discovered tasks to subsequent pages, allowing later exploration steps to skip redundant task discovery.

\noindent\textbf{Handle Fixed and Dynamic Elements Differently.} Most websites contain both fixed and dynamic GUI elements. Fixed elements are tied to the page template, stay in stable positions, and usually lead to consistent outcomes; for example, the search box on Reddit's home page always appears at the top and leads to the search results page. Dynamic elements instead depend on page content and may shift position, with outcomes varying by item; for example, different forum links on Reddit's forums page lead to different forums. AutoSurfer uses LLM reasoning to categorize discovered elements into fixed and dynamic sets, denoted as $\mathcal{S}_f$ and $\mathcal{S}_d$ (Algorithm~\ref{alg:exploration_task_tuple_generation}, Appendix \ref{appendix:categornizeelementsprompt}). It then attempts to cover all eligible fixed-element tasks, but samples a small representative subset of dynamic tasks, $\mathcal{S}_{dsamp} \subseteq \mathcal{S}_d$, for further exploration. For instance, on Reddit forums page, sampling 0--2 forum links is usually enough because different links lead to pages with similar structure even if their contents differ. This strategy preserves important functionality while reducing redundant exploration.

\noindent\textbf{Exclude Certain Tasks.} AutoSurfer excludes simple tasks involving authorization (e.g., login and logout), account creation or modification, payments, transient UI elements such as ads, pop-ups, and modals, and basic help elements such as tooltips. It relies on LLM reasoning to detect and filter these tasks during both initial and differential task discovery (Appendices \ref{appendix:taskdiscoveryprompt} and \ref{appendix:differentialtaskdiscoveryprompt}).

\noindent\textbf{Role of Queue.} The queue is central to AutoSurfer's systematic breadth-first exploration: it tracks visited pages to avoid redundant exploration, prioritizes high-level pages before deeper ones, and stores action traces that reconstruct the path to each discovered page. It also propagates knowledge from earlier pages to later ones, reducing redundant task discovery across the website.

\subsubsection{Complex Task Synthesis}
\label{sec:method:exploration:tasksynthesis}

AutoSurfer discovers complex tasks as it explores a website -- hence following a ``learn as you go'' strategy. It synthesizes complex tasks by combining actions from simple tasks that are logically connected during exploration. For example, AutoSurfer discovers `create a new forum' complex task in Reddit website by synthesizing the actions (1) `navigate to forums page' at home page, (2) `click on create forum button' at forums page, (3) `fill forum name' at create forum page, (4) `fill forum description' at the same page, and finally (5) `click on create button' at the same page. AutoSurfer supports the discovery of two types of complex tasks: (1) action-oriented tasks and (2) information-seeking tasks.

\noindent\textbf{Detection of Action-oriented Complex Tasks.} AutoSurfer follows OS-Genesis's prompt-based strategy \citep{sun2025osgenesisautomatingguiagent} to identify action-oriented complex tasks during exploration. When it observes a logically connected sequence of simple tasks spanning at least three GUI actions, it applies OS-Genesis prompts to synthesize the task and evaluate its completeness on a 1--5 scale, retaining only task tuples with score at least 3; we denote each accepted tuple as $t_{ao}$ (Algorithm~\ref{alg:exploration_task_tuple_generation}). Unlike OS-Genesis, AutoSurfer also records the corresponding action trace (Figure \ref{fig:method:tasksynthesis}) so it can later guide task-tuple refinement (Section \ref{sec:method:refinement}).

\noindent\textbf{Information-seeking Complex Tasks.} Information-seeking tasks complement action-oriented ones by targeting useful information on a page that users naturally seek. For example, `find top 5 trending posts in technology subreddit' requires both navigation and LLM reasoning over page content. When AutoSurfer reaches a page, it synthesizes candidate information-seeking tasks, denoted as $\mathcal{S}_{in}$ (Algorithm~\ref{alg:exploration_task_tuple_generation}, Appendix \ref{appendix:informationseekingtaskprompt}), and evaluates each resulting task tuple $t_{in} \in \mathcal{S}_{in}$ with OS-Genesis's task evaluator prompt. Similar to action-oriented tasks, AutoSurfer records the action trace for each accepted information-seeking task tuple for later refinement.

\noindent\textbf{Resemble Human Behavior in Exploration and Task Synthesis.} AutoSurfer's exploration strategy resembles how a human would learn a new website. It starts from the home page and explores high-level pages before deeper ones, avoids repeating actions that were already tried on earlier pages, and examines most fixed elements while sampling only a few dynamic ones to capture content-specific functionality efficiently.



Once the full exploration is done, the complete set of exploration task tuples $T_{ex}$ (Algorithm~\ref{alg:exploration_task_tuple_generation}) is generated.

\refstepcounter{algorithm}\label{alg:exploration_task_tuple_generation}
\begin{tcolorbox}[colback=gray!3,colframe=gray!50,boxrule=0.5pt,arc=2pt,left=6pt,right=6pt,top=6pt,bottom=6pt,breakable]
{\footnotesize\ttfamily
\textbf{Algorithm \thealgorithm: Exploration task tuple generation by AutoSurfer}\\
\textbf{Input:} seed page $p_0$; max complex tasks $k$\\
\textbf{Output:} explored complex task tuples $T_{ex}$\\
\\
1. $Q \leftarrow \textsc{Initialize}(p_0,\;[\;])$; \ $T_{ex} \leftarrow \{\}$\\
2. \textbf{while} $Q$ is not empty:\\
3. \ \ \ $(p,\;trace) \leftarrow Q.\textsc{Dequeue}()$\\
4. \ \ \ \textsc{Navigate}$(p,\;trace,\;Q,\;T_{ex},\;k)$\\
5. \ \ \ \textbf{if} $|T_{ex}| \geq k$: \textbf{break} \ \textit{// required task tuples are discovered}\\
6. \textbf{return} $T_{ex}$\\
\\
\textbf{Procedure} \textsc{Navigate}$(p,\;trace,\;Q,\;T_{ex},\;k)$\\
\\
1. \ $o \leftarrow \textsc{GetObservation}(p)$\\
2. \ $\mathcal{S}_{init} \leftarrow \textsc{GetSimpleTasksLLM}(o)$\\
3. \ $(\mathcal{S}_f,\;\mathcal{S}_d) \leftarrow \textsc{CategorizeFixedAndDynamicLLM}(o,\;\mathcal{S}_{init})$\\
4. \ $\mathcal{S}_{dsamp} \leftarrow \textsc{SampleDynamic}(\mathcal{S}_d)$\\
5. \ $\mathcal{S}_{in} \leftarrow \textsc{GetInfoSeekingTasksLLM}(o)$\\
6. \ $\mathcal{S} \leftarrow \mathcal{S}_f \cup \mathcal{S}_{dsamp} \cup \mathcal{S}_{in}$\\
7. \ \textbf{while} $\mathcal{S}$ is not empty:\\
8. \ \ \ $s \leftarrow \mathcal{S}.\textsc{Dequeue}()$\\
9. \ \ \ \textsc{GotoPage}$(p)$\\
10. \ \ \textsc{ExecuteAction}$(s)$\\
11. \ \ $o_{new} \leftarrow \textsc{GetObservation}(p)$\\
12. \ \ \textbf{if} \textsc{IsInfoSeekingTask}$(s)$:\\
13. \ \ \ \ $t_{in} \leftarrow \textsc{SynthesizeAndEvaluateComplexTaskLLM}(trace,\;s)$\\
14. \ \ \ \ \textbf{if} $t_{in} \neq \texttt{null}$: $T_{ex}.\textsc{Append}(t_{in})$; \ \textbf{if} $|T_{ex}| \geq k$: \textbf{return}\\
15. \ \ \textbf{else if} \textsc{IsNewPageDetected}$(o,\;o_{new})$:\\
16. \ \ \ \ $p_{new} \leftarrow \textsc{GetPage}(o_{new})$\\
17. \ \ \ \ \textbf{if not} $Q.\textsc{IsExplored}(p_{new})$:\\
18. \ \ \ \ \ \ $Q.\textsc{Enqueue}(p_{new},\;trace \,\|\, s)$\\
19. \ \ \ \ \ \ $t_{ao} \leftarrow \textsc{SynthesizeAndEvaluateComplexTaskLLM}(trace,\;s)$\\
20. \ \ \ \ \ \ \textbf{if} $t_{ao} \neq \texttt{null}$: $T_{ex}.\textsc{Append}(t_{ao})$; \ \textbf{if} $|T_{ex}| \geq k$: \textbf{return}\\
21. \ \ \textbf{else:} \ \textit{// same page}\\
22. \ \ \ \ \textbf{if} \textsc{ModerateVisualChange}$(o,\;o_{new})$:\\
23. \ \ \ \ \ \ $\mathcal{S}_{new} \leftarrow \textsc{GetNewlyVisibleSimpleTasksLLM}(o,\;o_{new})$\\
24. \ \ \ \ \ \ $\mathcal{S}.\textsc{Extend}(\mathcal{S}_{new})$\\
}
\end{tcolorbox}

\subsection{Task Tuple Refinement}
\label{sec:method:refinement}

Initial task tuples discovered during exploration (Sections \ref{sec:method:exploration:traversal} and \ref{sec:method:exploration:tasksynthesis}) are treated as non-refined because they may contain hallucinated or weakly grounded task descriptions, redundant or sub-optimal action sequences, and incomplete action coverage due to the exploratory policy. For example, exploration may omit actions such as scrolling because it often samples only the initially visible content and a few representative dynamic elements without needing to scroll.

To improve these tuples, AutoSurfer passes each explored task tuple to a web agent (we use SynthAgent's web agent) augmented with exploration traces as hints, and obtains a refined tuple $t_{ref} = f_{wa}(t_{ex})$, where $t_{ex} = (td_{ex}, \{o_{ex_{1}}, a_{ex_{1}}, o_{ex_{2}}, a_{ex_{2}}, ..., o_{ex_{n-1}}, a_{ex_{n-1}}, o_{ex_{n}}\})$ and $t_{ref} = (td_{ref}, \{o_{ref_{1}}, a_{ref_{1}}, o_{ref_{2}}, a_{ref_{2}}, ..., o_{ref_{m-1}}, a_{ref_{m-1}}, o_{ref_{m}}\})$. During execution, the agent simultaneously improves the task description and generates realistic trajectory for each task tuple. AutoSurfer discards a task tuple if the agent determines a task failure or if the agent exceeds 30 steps without completing the task.

\subsection{Model Fine Tuning}
\label{sec:method:finetuning}

Once AutoSurfer computes refined task tuples, it converts them into SFT data with the goal of training an LLM. The data supervises the next grounded GUI action from task description, current observation, and recent action history. The data is serialized into a conversational ShareGPT format, where the user message includes the task description, page screenshot, accessibility tree and up to three prior actions, and the assistant message includes the next action and the underlying reasoning.

Formally, each observation-action pair in a refined tuple $t_{ref}$ becomes a supervised example $g_{t_j} = (td_{ref}, f_{user}(o_j,a_{j-3},a_{j-2},a_{j-1}), f_{assistant}(a_j))$, and the final observation $o_m$ is paired with a dummy terminating action $a_m$. Thus, each refined tuple yields $G_t = \{g_{t_1}, g_{t_2}, ..., g_{t_m}\}$, and the full SFT dataset is $G = \bigcup_{t_{ref} \in T_{ref}} G_t$. As a proof of concept, we fine-tune multimodal Qwen models (Section~\ref{sec:experiments:models}) using the same hyper-parameters as SynthAgent \citep{wang2026adaptingwebagentssynthetic}.


\section{Datasets and Models}
\label{sec:experiments}

\subsection{Benchmark Datasets}
\label{sec:experiments:datasets}

We evaluate AutoSurfer on WebArena \citep{zhou2024webarenarealisticwebenvironment}, which contains 812 tasks across six docker-deployable realistic websites: CMS, Gitlab, Maps, Reddit, Shopping, and Wikipedia. We consider only single-website tasks because we mainly fine-tune one web agent per website. We exclude Wikipedia since its tasks require navigation to other websites and Maps since it currently has an internal issue. For efficiency, we evaluate one task per template, resulting in 57, 56, 26, and 55 tasks for CMS, Gitlab, Reddit, and Shopping, respectively.

\subsection{LLM Models}
\label{sec:experiments:models}

We use GPT-4.1 \citep{openai_gpt41_2025} for exploration, task synthesis, and refinement to ensure high-quality task tuples. For downstream fine-tuning and evaluation, we use the smaller open-source Qwen2.5-VL-7B-Instruct (or Qwen2.5) \citep{bai2025qwen25vltechnicalreport} and Qwen3-VL-8B-Instruct (or Qwen3) \citep{yang2025qwen3technicalreport}, which enables efficient adaptation while remaining strong enough to assess data quality. We keep the same fine-tuning and evaluation settings as SynthAgent \citep{wang2026adaptingwebagentssynthetic} for fair comparison.






\section{Results and Discussions}
\label{sec:results}

\subsection{Main Results from WebArena Benchmark}
\label{sec:results:accuracy}

We run AutoSurfer's exploration and refinement pipeline on each WebArena website to obtain refined complex task tuples and convert them into SFT data. In the first experiment, we fine-tune Qwen2.5 separately on each website's SFT data, resulting in four website-specific models; we denote this setting as AutoSurfer\_wllm. In the second experiment, we fine-tune Qwen2.5 on the combined SFT data from all four websites, resulting in a single model; we denote this setting as AutoSurfer\_webarena. We evaluate both AutoSurfer\_wllm and AutoSurfer\_webarena on the WebArena benchmark and compare them with other web agents. As shown in Table \ref{tab:results_without_map}, both SFT variants of AutoSurfer outperform the competing web agents in overall accuracy as well as in per-website accuracy. The performance of AutoSurfer\_webarena and AutoSurfer\_wllm is very similar, indicating that AutoSurfer can generate sufficiently high-quality SFT data even for a single website. In other words, AutoSurfer can reliably model a website without relying on data from other websites. To the best of our knowledge, AutoSurfer is the first approach that can train a wLLM with this level of accuracy through SFT.

In our analysis, we do not directly compare AutoSurfer with the GPT-4.1 baseline because GPT-4.1 is substantially larger than the other models considered in this paper. Nevertheless, the results suggest that AutoSurfer performs reasonably close to the GPT-4.1 baseline, which provides strong evidence for the quality of the data generated by AutoSurfer and the effectiveness of its fine-tuning strategy.

\begin{table}[h]
\centering
\begin{tabular}{lcccccc}
\hline
Method & Base  & Shopping & CMS & Reddit & Gitlab & Overall \\
 & model &  &  &  &  &  \\
\hline
GPT-4.1 baseline & GPT-4.1 & 30.91 & 24.56 & 15.38 & 26.79 & 25.77 \\
\hline
Qwen2.5 baseline & Qwen2.5 & 13.71 & 8.24 & 9.43 & 6.18 & 9.36 \\
Explorer & Qwen2.5  & 10.91 & 3.51 & 0.00 & 1.82 & 4.64 \\
OS-Genesis & Qwen2.5  & 14.55 & 10.53 & 11.54 & 16.07 & 13.40 \\
SynthAgent & Qwen2.5  & 20.00 & 21.05 & 15.38 & 19.64 & 19.59 \\
AutoSurfer\_wllm & Qwen2.5  & \textbf{23.64} & 22.81 & \textbf{34.62} & 19.64 & 23.71 \\
AutoSurfer\_webarena & Qwen2.5  & 20.00 & \textbf{24.56} & 26.92 & \textbf{26.79} & \textbf{24.23} \\
\hline
\end{tabular}
\caption{Performance comparison between AutoSurfer and other methods based on WebArena benchmark. 
Qwen2.5 baseline uses the original Qwen2.5 model, while other methods use the corresponding fine-tuned variant. AutoSurfer has two variants: AutoSurfer\_wllm and AutoSurfer\_webarena. The best performing method for each website and overall is highlighted in bold. GPT-4.1 baseline is excluded from the comparison as it is substantially larger than Qwen2.5, but its results show the upper bound performance by a large model.}
\label{tab:results_without_map}
\end{table}


\subsection{Results with a Different Model}

We fine-tune a different model, Qwen3 \citep{yang2025qwen3technicalreport}, on AutoSurfer-generated SFT data to examine whether the improvement remains consistent with Qwen2.5. In a small-scale experiment, we fine-tune Qwen3 using the same Reddit WebArena SFT data and compare the fine-tuned model against its original counterpart. AutoSurfer improves the model's performance by a large margin (30.77\% vs 3.85\%), which is consistent with the improvements observed when fine-tuning Qwen2.5 (Section \ref{sec:results:accuracy}). The low performance of the original model further suggests that even a stronger base model can still lack high-quality web interaction data needed for reliable task execution.
Due to time constraints, we run this experiment only for AutoSurfer. Nevertheless, we expect the relative trends for other methods to remain broadly consistent with the results observed using Qwen2.5.



\subsection{Task Diversity Analysis}


We analyze the diversity of synthesized complex tasks produced by each method to evaluate whether the method can discover a wide variety of tasks with high coverage and low redundancy. To quantify task diversity, we first embed each synthesized task title using the text-embedding-3-large model \citep{openai_text_embedding_large_3}. We then follow \cite{ethayarajh-2019-contextual} to compute the normalized average pairwise cosine similarity among the resulting task vectors. Finally, we convert the similarity score into a diversity score by subtracting it from 1, yielding a final score between 0 and 1, where a higher value indicates a more diverse task set. AutoSurfer achieves the highest diversity score of 0.7774, outperforming both OS-Genesis and SynthAgent (0.7459 for each). This result further supports AutoSurfer's superiority in comprehensive exploration and broader task-space coverage. We exclude Explorer from this analysis because it discovers fewer than two tasks per website on average, which is insufficient for a statistically meaningful diversity estimate.


\subsection{Exploration vs. Refined Trajectories}

To assess the importance of task-tuple refinement for SFT, we fine-tune Qwen2.5 using only exploration data and compare its performance with that of a model fine-tuned on refined data. In a small-scale experiment on the Reddit website from WebArena, the model trained on exploration data achieves an accuracy of 11.54\%, substantially lower than the 34.62\% achieved by the model trained on refined data (Section \ref{sec:results:accuracy}). This result indicates that exploration data alone does not provide sufficiently clean supervision for effective next-action learning, whereas the refinement stage recovers trajectories that are better suited for training.

One reason for this gap is that the exploration stage operates with a reduced action space relative to the refinement stage. For example, during exploration the agent does not scroll down the page, because much of the website's functionality can be discovered from the initially visible content. Similarly, when interacting with lists of dynamic elements, the exploration often needs to inspect only a small subset of items, making scrolling unnecessary. However, this restricted action space limits the model's ability to learn behaviors required for tasks that depend on omitted actions. In contrast, the refinement stage is centered on completing a specific task, which allows the agent to use the full action space, including scrolling when necessary. As a result, the refined trajectories provide more complete supervision and produce a model that is better equipped to execute real tasks.

The importance of refinement can be illustrated with a concrete example. AutoSurfer discovers the task ``How can I view all comments for the `New Spider-Man Image Leaked' post?'' by learning a sequence of functionalities: (1) search for the post, (2) edit the searched post, (3) comment on the searched post, (4) comment on a reply in the same post, and (5) analyze the resulting page. During refinement, the system determines that only searching and analyzing the post is necessary to complete the task, and therefore produces a more efficient trajectory by removing the redundant actions. At the same time, AutoSurfer may also discover tasks during exploration that are already correct and well optimized. For example, AutoSurfer discovers the task ``Sort the forum submissions list by selecting the 'New' option in the sorting menu to view the latest posts as soon as they appear'' by learning how to sort posts using the `New' option. In this case, the refinement step can simply execute the exploration trajectory without needing to modify it.


\subsection{Explanation behind Low Accuracies}


In this paper, we use a relatively simple LLM-based web agent developed by SynthAgent, whose performance depends primarily on the quality of the underlying model. In contrast, state-of-the-art web agents often incorporate stronger algorithms and system designs, so their performance reflects both model quality and agent design. We intentionally use this simpler agent because it enables a fairer comparison across methods based mainly on their corresponding fine-tuned models. As a result, the overall accuracies are low for all methods, since they largely reflect model capability rather than more advanced agentic components; although the absolute accuracies would likely improve with a stronger web agent, doing so would make it harder to isolate the contribution of the model from that of the agent.

\section{Conclusion}
\label{sec:conclusion}

In this paper, we presented AutoSurfer, a comprehensive web trajectory generator to build high-quality training data for web agents. By combining systematic breadth-first exploration, trajectory-grounded task synthesis, and trajectory-guided refinement, AutoSurfer more faithfully captures a website's action space and produces coherent task tuples for training wLLMs. AutoSurfer also yields strong per-website results and higher task diversity, suggesting that better exploration and refinement lead to more accurate and practical web agents. A promising future direction is to use AutoSurfer's refined trajectories as initialization for reinforcement learning with online interaction and task-completion rewards, enabling better adaptation to longer-horizon and cross-website tasks beyond static SFT.

\clearpage


\bibliography{bib}
\bibliographystyle{johd}

\appendix
\setcounter{figure}{0}
\renewcommand{\thefigure}{A\arabic{figure}}
\renewcommand{\theHfigure}{A\arabic{figure}}
\setcounter{table}{0}
\renewcommand{\thetable}{A\arabic{table}}
\renewcommand{\theHtable}{A\arabic{table}}

\section{Appendix}
\label{appendix}

\subsection{Related Works}
\label{sec:relatedwork}

\subsubsection{Web Agents}
\label{sec:relatedwork:webagents}

Web agents powered by multimodal LLMs have emerged as a promising paradigm for automating complex tasks in digital environments. A series of benchmarks have been developed to evaluate web agent capabilities, including MiniWoB++~\citep{Liu2018ReinforcementLO} for elementary browser tasks, Mind2Web~\citep{deng2023mindweb,online-mind2web} for cross-domain web navigation across 137 real-world websites, and WebArena~\citep{koh2024visualwebarena} for end-to-end evaluation in realistic, self-hostable web environments spanning e-commerce, content management, forums, and code repositories. Building upon these benchmarks, numerous agent frameworks have been proposed, ranging from vision-grounded approaches such as SeeAct~\citep{zheng2024seeact} and WebVoyager~\citep{he2024webvoyager}, to hierarchical planning methods like Agent S~\citep{Agent-S,Agent-S2,Agent-S3} and native vision-based agents such as UI-TARS~\citep{qin2025ui}. Despite rapid progress, web agent accuracy on complex, multi-step tasks remains well below human performance, underscoring the need for higher-quality training data.

\subsubsection{Data Synthesis for Building Web Agents}
\label{sec:relatedwork:datasynthesis}

The development of capable web agents critically depends on the availability of large-scale, high-quality trajectory data. Early efforts such as Mind2Web~\citep{deng2023mindweb,online-mind2web} relied on human annotators to demonstrate web tasks, producing high-fidelity data but at prohibitive cost and limited scale.

To overcome these limitations, task-driven synthesis methods automate trajectory generation by first defining tasks and then executing them. AgentTrek~\citep{xu2025agenttrek} harvests tutorial-like texts from the web to construct task specifications, which a web agent then executes in real environments. FaraGen~\citep{awadallah2025fara7b} uses a multi-agent pipeline to propose tasks from seed URLs, solve them, and verify successful trajectories, producing 145K trajectories across 70K domains. However, both approaches operate at a coarse level---AgentTrek depends on external tutorial availability, and FaraGen's task proposals lack deep exploration of individual website structures.

A more recent interaction-driven paradigm reverses the pipeline by first exploring environments and then deriving tasks. OS-Genesis~\citep{sun2025osgenesisautomatingguiagent} performs random step-wise interactions with GUIs and retrospectively synthesizes task instructions via \textit{reverse task synthesis}. Explorer~\citep{pahuja2025explorer} uses a multi-agent pipeline to generate 94K trajectories from 49K URLs, though it averages fewer than two trajectories per website, indicating limited depth per site. SynthAgent~\citep{wang2026adaptingwebagentssynthetic} improves data quality through categorized exploration and dual refinement of both tasks and trajectories. While these methods have advanced the state of the art, they share common limitations: random or shallow exploration strategies that miss hidden functionalities (e.g., multi-level menus and dropdowns), no explicit distinction between fixed and dynamic GUI elements, and a loose coupling between task synthesis and trajectory generation that can lead to incoherent task-trajectory pairs.

AutoSurfer addresses these gaps with a systematic breadth-first traversal that mirrors human exploration behavior, recursive expansion of hierarchical menu structures, differentiated handling of fixed versus dynamic elements, and tight coupling of task synthesis with trajectory generation to ensure coherent task-trajectory pairs.

\begin{figure}[t]
    \centering
    \includegraphics[width=0.9\linewidth]{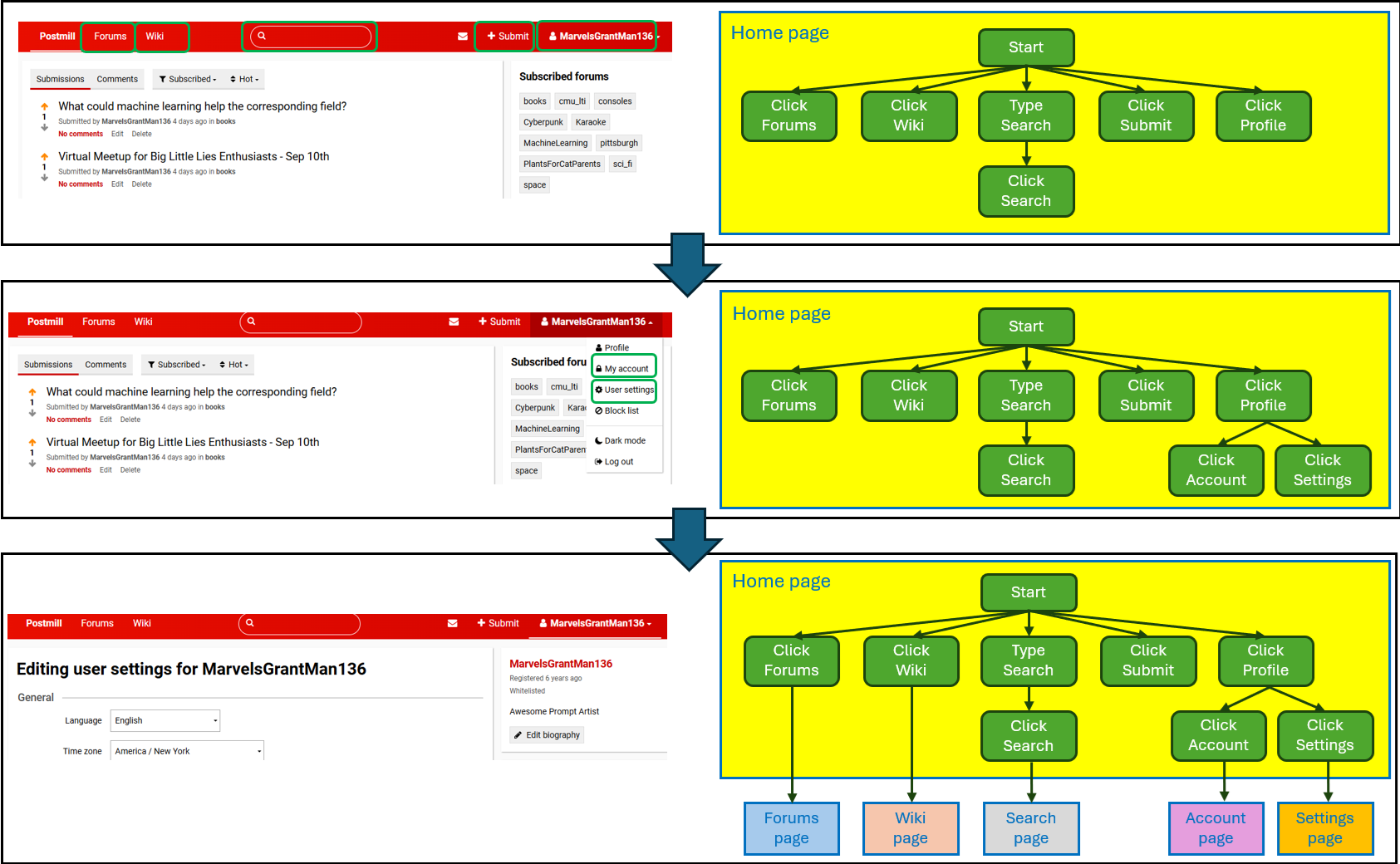}
    \caption{Illustration of AutoSurfer's breadth-first-style traversal in exploration and simple task discovery. In the top panel, AutoSurfer explores the home page, discovers simple tasks, such as `navigate to forums page', `perform search', `expand user profile', and constructs an exploration graph. Then AutoSurfer executes each discovered simple task to (a) discover new pages, or (b) discover new simple tasks from the same page, or (c) identify a completed complex task. In the middle panel, AutoSurfer expands the user profile menu to discover new simple tasks, such as, `navigate to user settings'. In the bottom panel, AutoSurfer discovers new pages by clicking on Forums link, Wiki link, account link and settings link, and performing search operation at the search box. AutoSurfer recursively explores other pages and performs similar operation until a termination condition is met.}
    \label{fig:method:traversal}
\end{figure}

\subsection{Prompts}

\subsubsection{Prompt to discover simple tasks from a web page}
\label{appendix:taskdiscoveryprompt}

\begin{promptbox}
{\small\sffamily
You are an intelligent agent capable of analyzing web pages and identifying short tasks based on screenshots and UI elements.

You are given:
\begin{itemize}[nosep,leftmargin=*]
  \item A screenshot of a web page with visible HTML elements annotated with red/yellow circles and numbers.
  \item The simplified HTML of the screenshot, where each HTML element has an ``id'' attribute that corresponds to the red/yellow circle number in the screenshot.
  \item A list of already observed short tasks with associated HTML elements.
\end{itemize}

Your goal is to exhaustively analyze the screenshot and HTML elements to identify all possible short tasks that can be performed on the web page.

Each short task is composed of a title and a sequence of UI actions required to accomplish that task.
Here are the supported UI action types:
\begin{itemize}[nosep,leftmargin=*]
  \item \texttt{goto}: Go to the specified URL. The action has one input which is the URL.
  \item \texttt{click}: Click on the specified HTML element. The action does not have any input.
  \item \texttt{fill}: Fill the specified HTML element with text input. The action has one input which is the text to fill.
  \item \texttt{press}: Press a key on the specified HTML element. The action has one input which is the key to press.
  \item \texttt{selectoption}: Select an option from a dropdown. The action has one input which is the option to select.
  \item \texttt{check}: Check a checkbox. The action does not have any input.
  \item \texttt{uncheck}: Uncheck a checkbox. The action does not have any input.
  \item \texttt{stop}: Stop the current action. The action does not have any input.
\end{itemize}

Instructions:
\begin{enumerate}[nosep,leftmargin=*]
  \item Examine every annotated element:
    \begin{itemize}[nosep,leftmargin=*]
      \item For each HTML element with an ``id'' attribute, determine if it can be used to perform a short task.
      \item Include elements that are viewed as pictures or icons.
    \end{itemize}
  \item Generate diverse short tasks:
    \begin{itemize}[nosep,leftmargin=*]
      \item Identify as many short tasks as possible as long as they can perform meaningful actions on the web page.
      \item First, identify short tasks that cover panel/header/sidebar elements, typically located at the top or left or right side of the screenshot.
      \item Second, identify short tasks for expandable UI elements, such as menus, dropdowns, and down-arrow buttons.
      \item Finally, identify short tasks for other elements.
      \item Include short tasks that involve various actions, such as navigation, clicking, filling forms, selecting options, checking/unchecking boxes, and pressing keys.
    \end{itemize}
  \item Respect action order in short tasks:
    \begin{itemize}[nosep,leftmargin=*]
      \item Some short tasks will require multiple actions to be completed in a specific order. As such, these actions should be collapsed into a single short task with an ordered action sequence.
      \item Example 1: ``Search for an item'' short task is completed by (1) filling the search box and (2) pressing ENTER.
      \item Example 2: ``Create post'' short task is completed by (1) filling title, (2) filling content, (3) selecting post category, and (4) clicking the ``Create post'' button.
      \item Some other short tasks will be simple and require only a single action.
      \item Two independent actions should not appear in the same short task.
    \end{itemize}
  \item Appropriate usage of multiple actions in short tasks:
    \begin{itemize}[nosep,leftmargin=*]
      \item A short task can have multiple actions when they appear as a group in the UI to perform a specific function.
      \item When a short task contains one or more fill actions, ensure that the short task ends with a non-fill action (e.g., click, press).
      \item A help or tooltip or formatting hint element should not be part of multiple action short tasks.
      \item If the same action appears in multiple short tasks, keep the longer short task and remove the shorter one.
      \item Make sure that mandatory elements (typically marked with asterisks) are covered by the corresponding short task.
    \end{itemize}
  \item Use realistic inputs:
    \begin{itemize}[nosep,leftmargin=*]
      \item Predict meaningful input values based on the context of the screenshot and HTML.
      \item For a fill action, provide realistic text input that a user would typically enter in that field.
      \item Carefully examine whether an input field can contain space characters, special characters, or only numbers, and provide input accordingly.
    \end{itemize}
  \item Set \texttt{is\_allowed} appropriately:
    \begin{itemize}[nosep,leftmargin=*]
      \item Use false for short tasks that require login, logout, sign up, account creation/modification, payment, or operate on transient UI elements like ads, pop-ups, or modals.
      \item Use false if the short task is already present in the list of observed short tasks.
      \item However, account viewing is allowed. Use true for viewing account or profile.
    \end{itemize}
  \item Avoid some tasks and actions:
    \begin{itemize}[nosep,leftmargin=*]
      \item Avoid any read text actions.
      \item Avoid any short task that operates on date pickers.
    \end{itemize}
\end{enumerate}

You are provided the following information:
\begin{itemize}[nosep,leftmargin=*]
  \item The screenshot of the web page with visible HTML elements annotated with red/yellow circles and numbers. The screenshot is attached with this message.
  \item The simplified HTML of the screenshot: \texttt{\{\{HTML\}\}}
  \item The list of already observed short tasks with associated HTML elements: \texttt{\{\{PAST\_ACTION\_SEQUENCES\}\}}
\end{itemize}

Generate the list of short tasks complying the following JSON schema:
\begin{quote}
\footnotesize\ttfamily
\{\\
\hspace*{1em}"task\_list": [\\
\hspace*{2em}\{\\
\hspace*{3em}"title": <title of the task>,\\
\hspace*{3em}"is\_allowed": <true if the task is allowed, false otherwise>,\\
\hspace*{3em}"action\_sequence": [\\
\hspace*{4em}\{\\
\hspace*{5em}"element\_id": <Id of GUI element>,\\
\hspace*{5em}"text": <text of the GUI element>,\\
\hspace*{5em}"type": <type of the action: fill, click, goto, stop, ...>,\\
\hspace*{5em}"input": [<input value 1>, ...]\\
\hspace*{4em}\},\\
\hspace*{4em}...\\
\hspace*{3em}]\\
\hspace*{2em}\},\\
\hspace*{2em}...\\
\hspace*{1em}]\\
\}
\end{quote}
}
\end{promptbox}

\subsubsection{Prompt to discover newly uncovered simple tasks from a web page}
\label{appendix:differentialtaskdiscoveryprompt}

\begin{promptbox}
{\small\sffamily
You are an intelligent agent capable of analyzing web pages and identifying short tasks based on screenshots and UI elements.

You are given:
\begin{itemize}[nosep,leftmargin=*]
  \item Two screenshots of the web page. The first screenshot is before the action is performed. The second screenshot is after the action is performed that has newly visible HTML elements highlighted with red/yellow circles and numbers.
  \item The simplified HTML of the second screenshot, where each HTML element has an ``id'' attribute that corresponds to the red/yellow circle number in the screenshot.
\end{itemize}

Your goal is to exhaustively analyze the screenshot and HTML elements to identify all possible short tasks from the newly visible part of the second screenshot.

Each short task is composed of a title and a sequence of UI actions required to accomplish that task.
Here are the supported UI action types:
\begin{itemize}[nosep,leftmargin=*]
  \item \texttt{goto}: Go to the specified URL. The action has one input which is the URL.
  \item \texttt{click}: Click on the specified HTML element. The action does not have any input.
  \item \texttt{fill}: Fill the specified HTML element with text input. The action has one input which is the text to fill.
  \item \texttt{press}: Press a key on the specified HTML element. The action has one input which is the key to press.
  \item \texttt{selectoption}: Select an option from a dropdown. The action has one input which is the option to select.
  \item \texttt{check}: Check a checkbox. The action does not have any input.
  \item \texttt{uncheck}: Uncheck a checkbox. The action does not have any input.
  \item \texttt{stop}: Stop the current action. The action does not have any input.
\end{itemize}

Instructions:
\begin{enumerate}[nosep,leftmargin=*]
  \item Examine every annotated element from the second screenshot:
    \begin{itemize}[nosep,leftmargin=*]
      \item For each HTML element with an ``id'' attribute, determine if it can be used to perform a short task.
      \item Include elements that are viewed as pictures or icons.
    \end{itemize}
  \item Generate diverse short tasks:
    \begin{itemize}[nosep,leftmargin=*]
      \item Identify as many short tasks as possible as long as they can perform meaningful actions on the web page.
      \item First, identify short tasks that cover panel/header/sidebar elements, typically located at the top or left or right side of the screenshot.
      \item Second, identify short tasks for expandable UI elements, such as menus, dropdowns, and down-arrow buttons.
      \item Finally, identify short tasks for other elements.
      \item Include short tasks that involve various actions, such as navigation, clicking, filling forms, selecting options, checking/unchecking boxes, and pressing keys.
    \end{itemize}
  \item Respect action order in short tasks:
    \begin{itemize}[nosep,leftmargin=*]
      \item Some short tasks will require multiple actions to be completed in a specific order. As such, these actions should be collapsed into a single short task with an ordered action sequence.
      \item For example, ``Search for an item'' short task is completed by (1) filling the search box and (2) pressing ENTER.
      \item Some other short tasks will be simple and require only a single action. For example, ``Add to cart'' short task in a shopping website is performed by one action - (1) click on the button.
      \item Two independent actions should not appear in the same short task. For example, clicking on user profile and clicking on notifications are independent actions and should not be part of the same short task.
    \end{itemize}
  \item Use realistic inputs:
    \begin{itemize}[nosep,leftmargin=*]
      \item Predict meaningful input values based on the context of the screenshot and HTML.
    \end{itemize}
  \item Set \texttt{is\_allowed} appropriately:
    \begin{itemize}[nosep,leftmargin=*]
      \item Use false for short tasks that require login, logout, sign up, account creation/modification, payment, or operate on transient UI elements like ads, pop-ups, or modals.
      \item However, account viewing is allowed. Use true for viewing account or profile.
    \end{itemize}
  \item Avoid some tasks and actions:
    \begin{itemize}[nosep,leftmargin=*]
      \item Avoid any read text actions.
      \item Avoid any short task that operates on date pickers.
    \end{itemize}
\end{enumerate}

You are provided the following information:
\begin{itemize}[nosep,leftmargin=*]
  \item Two screenshots of the web page where first one is before the action and the second one is after the action. The screenshots are attached with this message.
  \item The simplified HTML of the second screenshot: \texttt{\{\{HTML\}\}}
\end{itemize}

Generate the list of short tasks complying the following JSON schema:
\begin{quote}
\footnotesize\ttfamily
\{\\
\hspace*{1em}"task\_list": [\\
\hspace*{2em}\{\\
\hspace*{3em}"title": <title of the task>,\\
\hspace*{3em}"is\_allowed": <true if the task is allowed, false otherwise>,\\
\hspace*{3em}"action\_sequence": [\\
\hspace*{4em}\{\\
\hspace*{5em}"element\_id": <Id of GUI element>,\\
\hspace*{5em}"text": <text of the GUI element>,\\
\hspace*{5em}"type": <type of the action: fill, click, goto, stop, ...>,\\
\hspace*{5em}"input": [<input value 1>, ...]\\
\hspace*{4em}\},\\
\hspace*{4em}...\\
\hspace*{3em}]\\
\hspace*{2em}\},\\
\hspace*{2em}...\\
\hspace*{1em}]\\
\}
\end{quote}
}
\end{promptbox}

\subsubsection{Prompt to detect visual change in a web page}
\label{appendix:visualchangeprompt}

\begin{promptbox}
{\small\sffamily
You are an intelligent agent capable of identifying whether an action on a web page causes a moderate visual change based on screenshots.

You are given two screenshots of a web page. The first one is before an action is performed, and the second one is after the action is performed. Your job is to identify whether the action caused a moderate visual change on the web page.

Your response should comply the following JSON schema:
\begin{quote}
\footnotesize\ttfamily
\{\\
\hspace*{1em}"reason": <brief explanation of why the page changed moderately>,\\
\hspace*{1em}"answer": <true if moderate visual change, false otherwise>\\
\}
\end{quote}
}
\end{promptbox}

\subsubsection{Prompt to categorize GUI elements in a web page}
\label{appendix:categornizeelementsprompt}

\begin{promptbox}
{\small\sffamily
You are an intelligent agent capable of analyzing the HTML and screenshot of a web page to identify and categorize short tasks based on UI elements.

You are given:
\begin{itemize}[nosep,leftmargin=*]
  \item A screenshot of a web page with visible HTML elements annotated with red/yellow circles and numbers.
  \item The simplified HTML of the screenshot, where each HTML element has an ``id'' attribute that corresponds to the red/yellow circle number in the screenshot.
  \item A list of identified short tasks with the annotated UI elements.
\end{itemize}

Your goal is to analyze the screenshot, HTML elements and short tasks (with annotated UI elements) and then categorize each short task into ``fixed'' or ``dynamic''.

You must follow the guidelines below:
\begin{enumerate}[nosep,leftmargin=*]
  \item An element is ``fixed'' if it has fixed position on the page and provides specific functionality. For example, ``search box'' at Amazon home page is ``fixed''.
  \item An element is ``dynamic'' if it does not have a fixed position on the page. Most commonly, ``dynamic'' elements are list of items, such as product lists and search results, providing similar functionalities. For example, ``product list'' at Amazon search results page is ``dynamic''.
  \item Assign a group number for each dynamic UI element. If a list of UI elements appear as a group then each UI element in that group should be assigned the same group number. For example, if there are 20 products from a search result in a shopping website, then all those 20 product UI elements should be assigned the same group number.
  \item There can be multiple groups of dynamic UI elements on the same page. As such, each group should have a different group number. For example, if there are 20 products from a search result and 10 recommended products on the same page, then the 20 products should be assigned one group number and the 10 recommended products should be assigned another group number.
  \item Short tasks in each dynamic group should perform similar actions/functions. Sometimes, there are UI elements that appear closer in a group but perform quite different actions/functions. In such cases, those UI elements should be assigned different group numbers. For example, in a list of product search results, there are links to product details and buttons to add products to cart. In this case, the product detail links and add to cart buttons should be assigned different group numbers even though they appear closer in a group.
  \item Usually following UI elements are not dynamic:
    \begin{itemize}[nosep,leftmargin=*]
      \item UI elements located at top or left or right header/panel/sidebar.
      \item Expandable UI elements, such as menus, dropdowns, and down-arrow buttons.
      \item UI elements appeared as a list of tabs.
    \end{itemize}
\end{enumerate}

You are provided the following information:
\begin{itemize}[nosep,leftmargin=*]
  \item The screenshot of the web page with visible HTML elements annotated with red/yellow circles and numbers. The screenshot is attached with this message.
  \item The simplified HTML of the screenshot: \texttt{\{\{HTML\}\}}
  \item The list of identified short tasks with annotated HTML elements: \texttt{\{\{SHORT\_TASKS\}\}}
\end{itemize}

Generate the list of categorized short tasks complying the following JSON schema:
\begin{quote}
\footnotesize\ttfamily
\{\\
\hspace*{1em}"task\_list": [\\
\hspace*{2em}\{\\
\hspace*{3em}"title": <title of the task>,\\
\hspace*{3em}"category": <fixed or dynamic>,\\
\hspace*{3em}"group\_number": <group number for dynamic tasks only>\\
\hspace*{2em}\},\\
\hspace*{2em}...\\
\hspace*{1em}]\\
\}
\end{quote}
}
\end{promptbox}

\subsubsection{Prompt to synthesize information-seeking complex tasks}
\label{appendix:informationseekingtaskprompt}

\begin{promptbox}
{\small\sffamily
You are an intelligent agent capable of analyzing a web page and identifying information that a user highly like to seek from the web page.

You are given:
\begin{itemize}[nosep,leftmargin=*]
  \item The screenshot of the web page.
  \item The simplified HTML of the web page: \texttt{\{\{HTML\}\}}
  \item Action history: \texttt{\{\{ACTION\_HISTORY\}\}}
\end{itemize}

Your goal is to analyze the screenshot, HTML element, and action history and then identify a list of potential asks that a user highly like to seek from the web page.

You must follow the principles below:
\begin{itemize}[nosep,leftmargin=*]
  \item Generate everything based on the screenshot, HTML element, and action history.
  \item Generate between 0 and 2 asks that represent natural information seeking behavior of a user.
  \item Don't generate any ask if a user is unlikely to seek any information from the web page.
\end{itemize}

Generate the list of asks complying the following JSON schema:
\begin{quote}
\footnotesize\ttfamily
\{\\
\hspace*{1em}"ask\_list": [\\
\hspace*{2em}\{\\
\hspace*{3em}"reason": <reason why the information is useful to seek>,\\
\hspace*{3em}"ask": <information-seeking question>\\
\hspace*{2em}\},\\
\hspace*{2em}...\\
\hspace*{1em}]\\
\}
\end{quote}
}
\end{promptbox}

\end{document}